\documentclass[]{article}
\usepackage{nips11submit_e,times}

\usepackage{graphicx} 
\usepackage{subfigure}
\usepackage{bbm}
\usepackage{amsmath}
\usepackage{amsthm}
\usepackage{amsfonts}
\usepackage{amssymb}
\usepackage{multirow}
\usepackage{wrapfig}
\usepackage{epstopdf}
\usepackage{natbib}

\usepackage{algorithm}
\usepackage{algorithmic}

\usepackage{hyperref}

\def\argmax{\mathop{\mathrm{argmax}}}

\newtheorem{theorem}{Theorem}

\newtheorem{definition}{Definition}

\title{Dynamic Batch Bayesian Optimization}

\author{
Javad Azimi\\
EECS, Oregon State University\\
\texttt{azimi@eecs.oregonstate.edu} \\
\And
Ali Jalali \\
ECE, University of Texas at Austin \\
\texttt{alij@mail.utexas.edu} \\
\AND
Xiaoli Fern \\
EECS, Oregon State University\\
\texttt{xfern@eecs.oregonstate.edu} \\
}
\nipsfinalcopy
\begin{document}

\maketitle

\begin{abstract}
Bayesian optimization (BO) algorithms try to optimize an unknown function that is  expensive to evaluate using minimum number of evaluations/experiments. Most of the proposed algorithms in BO are sequential, where only one experiment is selected at each iteration. This method can be time inefficient when each experiment takes a long time and more than one experiment can be ran concurrently. On the other hand, requesting a fix-sized batch of experiments at each iteration causes performance inefficiency in BO compared to the sequential policies. In this paper, we present an algorithm that asks a batch of experiments at each time step $t$ where the batch size $p_t$ is dynamically determined in each step.  Our algorithm is based on the observation that the sequence of experiments selected by the sequential policy can sometimes be almost independent from each other. Our algorithm identifies such scenarios and request those experiments at the same time without degrading the performance. We evaluate our proposed method using the \emph{Expected Improvement} policy and the results show substantial speedup with little impact on the performance in eight real and synthetic benchmarks.
\end{abstract}

\section{Introduction}

\emph{Bayesian Optimization}(BO) algorithms try to optimize an unknown function $f(\cdot)$ by requesting a set of experiments (evaluation of the function at requested points) when the function is costly to evaluate \cite{taxonomy, Brochu09}. In general, after selecting an experiment based on a selection criterion, we need to wait for the results of the experiment to select the next experiment (based on a better prior). This is commonly referred to as a \emph{sequential policy}. The sequential framework is not efficient in applications where running an experiment is time consuming and we have the ability to run multiple experiments concurrently.

Recently, Azimi \emph{et. al.} \cite{azimi10} introduced a \emph{batch} BO approach that selects a batch of experiments at each iteration that best approximates the behavior of a given sequential heuristic such as \emph{Expected Improvement} (EI). Their results show that batch selection in general performs worse than a sequential policy, especially when the total number of experiments is small. This motivates us to introduce a \emph{dynamic batch} BO approach that selects a varying number of experiments at each time step. Our goal is to select experiments that are likely to be selected by a given sequential policy. Specifically, at each step $t$ , given a sequential policy (EI in this paper), we look for the possibility that the next $p_t\geq 1$ samples $x_1, x_2,...,x_{p_t}$ selected by EI are approximately independent from each other. That is, EI is likely to select example $x_i$ in the $i$-th step regardless of the outcome of the previous $i-1$ experiments. Upon finding such a set of independent experiments, we can ask for those experiments in batch and hence speedup our experiment process. Note that we might not be able to ask for more than one experiment at some iterations, since the next experiment might strongly depend on the currently selected/running experiments.

\section{Gaussian Process (GP)} 
Any BO algorithm consists of two main components, 1) the model of the unknown function which generates a posterior over the outputs of unobserved points in the input space, and 2) a \emph{selection criterion}, which chooses an experiment at each iteration based on  available prior. We use \emph{Gaussian Process} (GP)\cite{Rasmussen06} as our primary model to build the posterior over the outcome values given our observations $\mathcal{O}=(\boldsymbol{x}_\mathcal{O},\boldsymbol{y}_\mathcal{O})$ where $\boldsymbol{x}_\mathcal{O}=\{x_1,x_2,\cdot\cdot\cdot,x_n\}$ and  $\boldsymbol{y}_\mathcal{O}=\{y_1,y_2,\cdot\cdot\cdot y_n\}$ such that $y_j=f(x_j)$ and $f(\cdot)$ is our underlying function. 

For a new input point $x_i$, GP models the output $y_i$ as a normal random variable $y_i\sim\mathcal{N}(\mu_i, \sigma^2_i)$, with  $\mathbf{\mu}_i=k(x_i,\boldsymbol{x}_\mathcal{O})k(\boldsymbol{x}_\mathcal{O},\boldsymbol{x}_\mathcal{O})^{-1}\boldsymbol{y}_\mathcal{O}$ and
$\mathbf{\sigma}_i^{2}=k(x_i,x_i)-k(x_i,\boldsymbol{x}_\mathcal{O})k(\boldsymbol{x}_\mathcal{O},\boldsymbol{x}_\mathcal{O})^{-1} k(\boldsymbol{x}_\mathcal{O},x_i)$ where $k(\cdot,\cdot)$ is an arbitrary symmetric positive definite \emph{kernel function} to compute the coherence between any pair of elements. We use \emph{squared exponential} as our kernel function, $k(x,y)= \text{exp}( - \frac{1}{l}\parallel x - y \parallel ^2)$ where $\parallel \cdot \parallel $ is the vector $2$-norm and $l$ is the constant length scale parameter. Our approach is flexible to the choice of kernel function.

\begin{definition}
Let $\boldsymbol{x}^*=\left\{x_1^*,x_2^*,\cdot\cdot\cdot,x_m^*\right\} \in \mathcal{X}\setminus\boldsymbol{x}_\mathcal{O}$ be any unobserved set of points. Let $\boldsymbol{y}^*=\{y_1^*,y_2^*,\cdot\cdot\cdot y_m^*\}$ be their corresponding predictions obtained from the Gaussian Process model learned from existing observations, where $y^*_i|\mathcal{O}\;\sim\;\mathcal{N}(\mu_i^*,\sigma^{*2}_i)$. Define $\boldsymbol{\sigma^{*2}}=\left(\sigma^{*2}_1,\sigma^{*2}_2,\cdot,\cdot\cdot,\sigma^{*2}_m\right)$ and $\boldsymbol{\mu^*}=\left(\mu^*_1,\mu^*_2,\cdot,\cdot\cdot,\mu^*_m\right)$ and for any point $z\in\mathcal{X}\setminus\left\{\boldsymbol{x}_\mathcal{O}\cup \boldsymbol{x}^*\right\}$, let $p(y_z|z,\mathcal{O})\sim\mathcal{N}(\mu_z,\sigma^2_z)$ and $p(y_z|z,\mathcal{O},(\boldsymbol{x}^*,\boldsymbol{y}^*))\sim\mathcal{N}(\mu_z^*,\sigma^{*2}_z)$.
\end{definition}

Using GP as our primary model, the variance of any point $z$ depends only on the location of the observed points and is independent from the observation outputs. Therefore, we can easily evaluate the variance of any point $z$ after requesting experiments at any set of points $\boldsymbol{x}^*$ without knowing their observation outputs. The following theorem characterizes the variance of $z$ after sampling at $\boldsymbol{x}^*$.

\begin{theorem}  Let $\Delta^*(\sigma_z):= \sigma_z^2-\sigma^{*2}_z$, then
\begin{equation}
\Delta^*(\sigma_z)= \left(PA^{-1}B^T-k^*_z\right)m\left(PA^{-1}B^T-k^*_z\right)^T,
\vspace{-0.05in}
\end{equation}
where \small$B=k(\boldsymbol{x}^*,\boldsymbol{x}_\mathcal{O})$, $A=k(\boldsymbol{x}_\mathcal{O},\boldsymbol{x}_\mathcal{O})$, $P=k(z,\boldsymbol{x}_\mathcal{O})$, $m=(k(\boldsymbol{x}^*,\boldsymbol{x}^*)-BA^{-1}B^T)^{-1}$ and $k_z^*=k(z,\boldsymbol{x}^*)$.\normalsize
\label{theorem:varchange}
\end{theorem}

The expected output value $\mu_z^*$ of any point $z$, however, heavily depends on the outputs of points $\boldsymbol{x}^*$, which are not available. Below, we provide an upper-bound on the change of $\mu_z$ after sampling at $\boldsymbol{x}^*$.
\vspace{0.1in}
\begin{theorem}  Let $\Delta^*(\mu_z)=\mu_z-\mu^*_z$, then
\small\begin{equation}
\mathbb{E}_{\boldsymbol{y}^*}[|\Delta^*(\mu_z)|]\leq\left\|\left(PA^{-1}B^T-k^*_z\right)m\right\|_\infty  \sqrt{\frac{2}{\pi}}\|\boldsymbol{\sigma^{*}}\|_1,
\end{equation}\normalsize
\label{theorem:expchange}
where $\parallel \cdot \parallel$ is the vector norm.
\end{theorem}

Interestingly, the above stated upper-bound of $\mathbb{E}_{\boldsymbol{y}^*}[|\Delta^*(\mu_z)|]$ can be considered as a function of sum of $\boldsymbol \sigma^{*}$ and independent from the observation outputs. Therefore, $\mathbb{E}_{\boldsymbol{y}^*}[|\Delta^*(\mu_z)|]$ converges to zero as the number of observations increases since the variance decreases.

\section{Dynamic Batch Design for Bayesian Optimization}

In the sequential approach, we ask for only one experiment at each time step using a given policy $\pi$ (the selection criterion). Suppose we have the capability of running $n_b$ experiments in parallel, and we are limited by the total number of possible experiments $n_l$. At each time step $t$, the question is whether or not we can select $1< p_t\leq n_b$ samples to speed up the experimental procedure without loosing any performance comparing to the sequential policy $\pi$. In \cite{azimi10}, the authors tried to select a batch of $k$ examples by predicting the selected samples at the next $k$ steps by a given policy $\pi$ using Monte Carlo simulation. However, frequently the predicted set of experiments are not exactly identical to those selected by the sequential policy (e.g. EI)  especially when the batch size $k$ is large. In this section, we introduce an algorithm that selects $p_t\geq 1$ experiments at each iteration based on (but not restricted to) the EI policy, where the batch size $p_t$ is dynamically determined at each step. In particular, it will select more than one experiments when/if the next set of experiments to be selected by $\pi$ are believed to be independent from the outcome of the already selected experiments. Algorithm \ref{alg:SBEI} is our proposed solution to this problem. We focus on the problem of finding the maximizer of an unknown function for the rest of the paper, however, the result can be easily extended to minimization applications.

\begin{definition}
Expected Improvement (EI) \cite{Locatelli97} at any point $x$ with corresponding GP prediction $y=\mathcal{N}(\mu_x, \sigma_x^2)$ is defined to be
\begin{equation}
\label{eq:mei}
EI(x)= \left(-u\Phi(-u)+\phi(u)\right)\sigma_x,
\end{equation}
where, $u=(y_{max}-\mu_x)/\sigma_x$ and $\displaystyle y_{max}=\max_{y_i\in\boldsymbol{y}_\mathcal{O}}y_i$\normalsize. Also, $\Phi(\cdot)$ and $\phi(\cdot)$ represent standard Gaussian distribution and density functions respectively.
\end{definition}

In general, we have to predict the selected experiments by EI for a sequence of consecutive time steps. Clearly, we are certain about the first experiment $x_1^*$ and we are interested in the EI value of the other selected points after sampling at point $x_1^*$. Therefore, the goal is to predict the EI values of other points after sampling at any point $x_1^*$. After selecting the sample $x_1^*$, we can calculate the variance of the output of any point $z$ regardless of the output value of $x_1^*$; but, the expectation of the output corresponding to $z$ can highly depend on the output value of $x_1^*$. Therefore, we cannot evaluate the \emph{exact} EI of any point $z$ before running the real experiment and observing the output. However, we can calculate an \emph{upper bound} on the EI of the next experiments. Below, we will show that the EI upper bound of all points in the space can be computed by simply setting the output value of $x^*_1$ to $M$, where $M$ is the maximum possible output value (which is easy to estimate or provide /given in many applications).

\begin{theorem}
Fixing the variance $\sigma_x^2\;$, EI is an increasing function of $\mu_x$.
\end{theorem}

As a consequence of this theorem, by setting the output value of $x_1^*$ to the maximum $M$, we set an upper bound for the expectation of any point $z$ that its mean and variance is changed after sampling at point $x_1^*$. Therefore, the EI of any point $z$ based on the current observation $\mathcal {O}\cup (x_1^*,M)$ can be considered as the EI upper bound value after observing the output value of $x_1^*$. Thus, we select the next sample $z$ based on current observation $\mathcal {O}\cup (x_1^*,M)$. If the next selected sample $z$ satisfies $\mathbb{E}[|\Delta^*(\mu_z)|]<\epsilon$, then the expectation of its output changes at most by $\epsilon$. For small values of $\epsilon$, we can consider its expectation as constant before and after sampling $x^*_1$ and hence $z$ is independent of $x_1^*$, experiment-wise. This entails that the point $z$ is likely to be the next selected experiment by EI policy since its EI value is more than the EI upper bound of the points affected by sampling $x^*_1$.

\begin{algorithm} 
\caption{Dynamic Batch Expected Improvement Algorithm}
\label{alg:SBEI}
{\bf Input:} Total budget of experiments ($n_l$), maximum batch size ($n_b$), the maximum observation value ($M$), current observation $\mathcal{O}=(\boldsymbol{x}_\mathcal{O}$,$\boldsymbol{y}_\mathcal{O}$) and stopping threshold $\epsilon$.
\small{
\begin{algorithmic}

		\WHILE {$n_l>0$}
			\STATE $\displaystyle x^*_1\gets\arg\;\max_{x\in \mathcal{X}}\;\; \mbox{EI}(x|\mathcal{O})$.
			\STATE $\mathcal{A}\gets(x^*_1,M)$, $\quad n_l\gets n_l-1$.\vspace{0.2cm}
			\STATE $\displaystyle z\gets\arg\;\max_{x\in \mathcal{X}}\;\; \mbox{EI}(x|\mathcal{O}\cup\mathcal{A})$.\vspace{0.1cm}
			\WHILE{$(n_l>0)$ and $(|\mathcal{A}|< n_b)$ and $\left(\mathbb{E}\left[|\Delta^*(\mu_z)|\right]\leq \epsilon\right)$}\vspace{0.1cm}
				\STATE $\mathcal{A}\gets\mathcal{A}\cup(z,M)$, $\quad n_l\gets n_l-1$.\vspace{0.1cm}
				\STATE $\displaystyle z\gets\arg\;\max_{x\in \mathcal{X}}\;\; \mbox{EI}(x|\mathcal{O}\cup\mathcal{A})$.
			\ENDWHILE \vspace{0.1cm}
			\STATE $\boldsymbol{y}^*\gets  \text{RunExperiment}(\boldsymbol{x}^*_\mathcal{A})$
			\STATE $\mathcal{O}\gets \mathcal{O}\cup (\boldsymbol{x}^*_\mathcal{A},\boldsymbol{y}^*)$
		\ENDWHILE
	  \STATE  \textbf{return}  $\max(\boldsymbol{y}_\mathcal{O})$
\end{algorithmic}}
\end{algorithm}

This algorithm is based on two main observations: (1) The selected sample $z$ is far \emph{enough} from $x_1^*$ and its expectation does not change after sampling at $x_1^*$; (2) The EI upper bound of affected samples after sampling at $x_1^*$ is less than the EI of any point $z$. Therefore, the point $z$ would likely be the next selected point for small value of $\epsilon$. This procedure is repeated until the next sample has $\mathbb{E}[|\Delta^*(\mu_z)|]>\epsilon$  indicating that we cannot find the next independent point.

\subsection{Other Sequential Approaches} 
The proposed approach can be extended to other sequential policies such as Maximum Mean (MM), Maximum Upper Interval (MUI) and Maximum Probability of Improvement (MPI) \cite{taxonomy}; however, the rate of the speedup would be significantly different among these approaches. For example, in MM with the objective function $x^*=\arg\max_{x\in\mathcal{X}}\,\mu_{x|\mathcal{O}}$, we would not have any speedup since we are only interested in the point with the highest mean. Therefore, by setting the output of the selected point $x^*$ to the highest possible value, the next selected sample would be one of the closest samples to $x^*$. For MUI policy with the objective function $x^*=\argmax_{x\in\mathcal{X}}\,\mu_{x|\mathcal{O}}+1.96\,\sigma_{x|\mathcal{O}}$, we expect a high rate of speedup since it is dominated by variance that is completely known by sampling at $x^*$. For MPI with the objective function \small$x^*=\max_{x\in\mathcal{X}}\, \Phi\left(\frac{\mu_{x|\mathcal{O}}-(1+\alpha)y_{\max}}{\sigma_{x|\mathcal{O}}}\right)$\normalsize, the speedup varies by varying the parameter $\alpha$.

\renewcommand{\arraystretch}{1.35}
\begin{table}[t]
\begin{center}
\caption{Benchmark Functions}\label{tab:functions}
\footnotesize{
\begin{tabular}{|l|c||l|c|}  \hline
\multirow{2}{*}{Cosines(2)} & \footnotesize $1\!-(u^2\!+v^2\!-0.3\cos(3\pi u)\!-0.3\cos(3\pi v))$ & \multirow{2}{*}{Rosenbrock(2)} & \multirow{2}{*}{\footnotesize$10\!-\!100(y\!-x^2)^2\!\!-\!(1\!-x)^2$}\\
&\footnotesize $u=1.6x-0.5, v=1.6y-0.5$ & & \\\hline
\multirow{2}{*}{Hartman(3,6)} & \footnotesize$\sum_{i=1}^4 \alpha_i \exp\left(-\!\sum_{j=1}^d A_{ij}(x_j-P_{ij})^2\right)$ & \multirow{2}{*}{Michalewicz(5)}& \multirow{2}{*}{\footnotesize$-\sum_{i=1}^{5}\sin(x_i)\sin\left(\frac{i\, x_i^2}{\pi}\right)^{\!\!20}$}\\
 & \footnotesize $\alpha_{1\times 4}, \; A_{4 \times d},\; P_{4\times d}$ are constants &  & \\\hline
Shekel(4)& \multicolumn{3}{c|}{ \footnotesize$\sum_{i=1}^{10}\frac{1}{\alpha_i+\Sigma_{j=1}4(x_j-A_{ji})^2}$ \qquad $\alpha_{1\times 10}, \; A_{4 \times 10}$ are constants}\\ \hline
\end{tabular}}
\end{center}
\end{table}

\begin{table}[t]
\caption{Benchmark Performance}\label{table:perf}
\begin{center}
\footnotesize{
\begin{tabular}{|l|c|c|c|c|c|c|c|c|}
\hline
 & \bf Cosines&\bf Hydrog&\bf FC&\bf Rosen&\bf Hartman 3&\bf Michal&\bf Shekel& \bf Hartman 6 \\
\hline
\bf Sequential&	$0.145$&	$0.026$&	$0.155$&	$0.005$&	$0.033$&	$0.369$&	$0.340$&	$0.222$\\
\hline
\bf Semi($M$)&	$0.148$&	$0.026$&	$0.151$&	$0.005$&	$0.036$&	$0.379$	&$0.335$&	$0.220$  \\
\bf Speedup($M$)& $6.8\%$&	$9.3\%$&	$10.2\%$&	$16.3\%$&	$10.1\%$&	$6.0\%$&	$2.1\%$&	$4.5\%$\\
\hline
\bf Semi($\alpha=0.1$) &$0.147$&	$0.027$&	$0.160$&	$0.006$ &	$0.034$&	$0.375$&	$0.345$&	$0.233$   \\
\bf Speedup($\alpha=0.1$)&	16.23\%&	$11.4\%$&	$11.7\%$&	$17.8\%$&	$18.4\%$&	$16.7\%$&	$17.4\%$&	$13.77\%$\\
\hline
\end{tabular}}
\end{center}
\end{table}

\section{Experimental Results}

As we mentioned earlier, we use GP to build a posterior over our unknown function $f(\cdot)$. Our GP is based on zero-mean prior and \emph{squared exponential} as covariance kernel function with kernel width $l=0.01\Sigma_{i=1}^d l_i$ with $l_i$ being the length of the $i^{th}$ dimension \cite{azimi10}. We consider six well-known synthetic benchmarks: Cosines and Rosenbrock \cite{Anderson00,Brunato06} over $[0,1]^2$, Hartman(i) \cite{GOP78} over $[0,1]^i$ for $i=3,6$, Shekel \cite{GOP78} over $[3,6]^4$ and Michalewicz \cite{Michal94} over $[0,\pi]^5$. The mathematical formulation of these functions are listed in Table \ref{tab:functions}. We also evaluate our approach on two real benchmarks Fuel Cell and Hydrogen over $[0,1]^2$. More details about these benchmarks can be found in \cite{azimi10}.

We run our algorithm on each benchmark $100$ times and report the average regret; $M-\max_{y_i \in \boldsymbol{y}_{\mathcal{O}}}y_i$. In each run, the algorithm starts with $5$ initial random points for $2,3$-dimensional benchmarks and $20$ initial random points for higher dimensional benchmarks. The number of experiments $n_l$ is set to $20$ for $2,3$-dimensional and $60$ for the higher dimensional benchmarks and the maximum batch size at each iteration, $n_b$, is set to $5$.  The parameter $\epsilon$ is set to $0.02$ for $2,3$-dimensional and $0.2$ for higher dimensional benchmarks.

Table \ref{table:perf} shows the performance of our algorithm on each benchmark along with the performance of the sequential policy. There are two different sets of results for each benchmark with different values of $M$ and $\alpha$. Since setting the value of any selected sample $z$ to a fixed value $M$ does not match the reality, we consider an improvement over the best current observed output, $y_{\max}$, as a surrogate to the fixed $M$; i.e., in each step, instead of $M$, we use the value $(1+\alpha)y_{\max}$. In this experiment, we set the value of $\alpha$ to $0.1$ which means we expect $10\%$ improvement over the best current observation after each sampling.

To illustrate the speedup over the sequential policy, we calculate the speedup as the percentage of the samples in the whole experiment that are selected to be run in parallel in the batch mode. In general, if we exhaust $n_l$ samples in $T$ steps, the speed up is calculated as $(n_l-T)/n_l$. Clearly, the maximum speedup in our setting is $0.8$ that can only happen if we select $5$ experiments at each and every time steps. The results show that, in both settings, we perform very close to the sequential policy and we achieve the maximum speedup of $16.3\%$ with fixed $M$ in Rosenbrock and $18.4\%$ with varying $M$ by $\alpha=0.1$ for Hartman$3$.

\begin{figure}
\begin{center}
\begin{tabular}{@{}c @{}}
\includegraphics[width=3 in,height=2.5in]{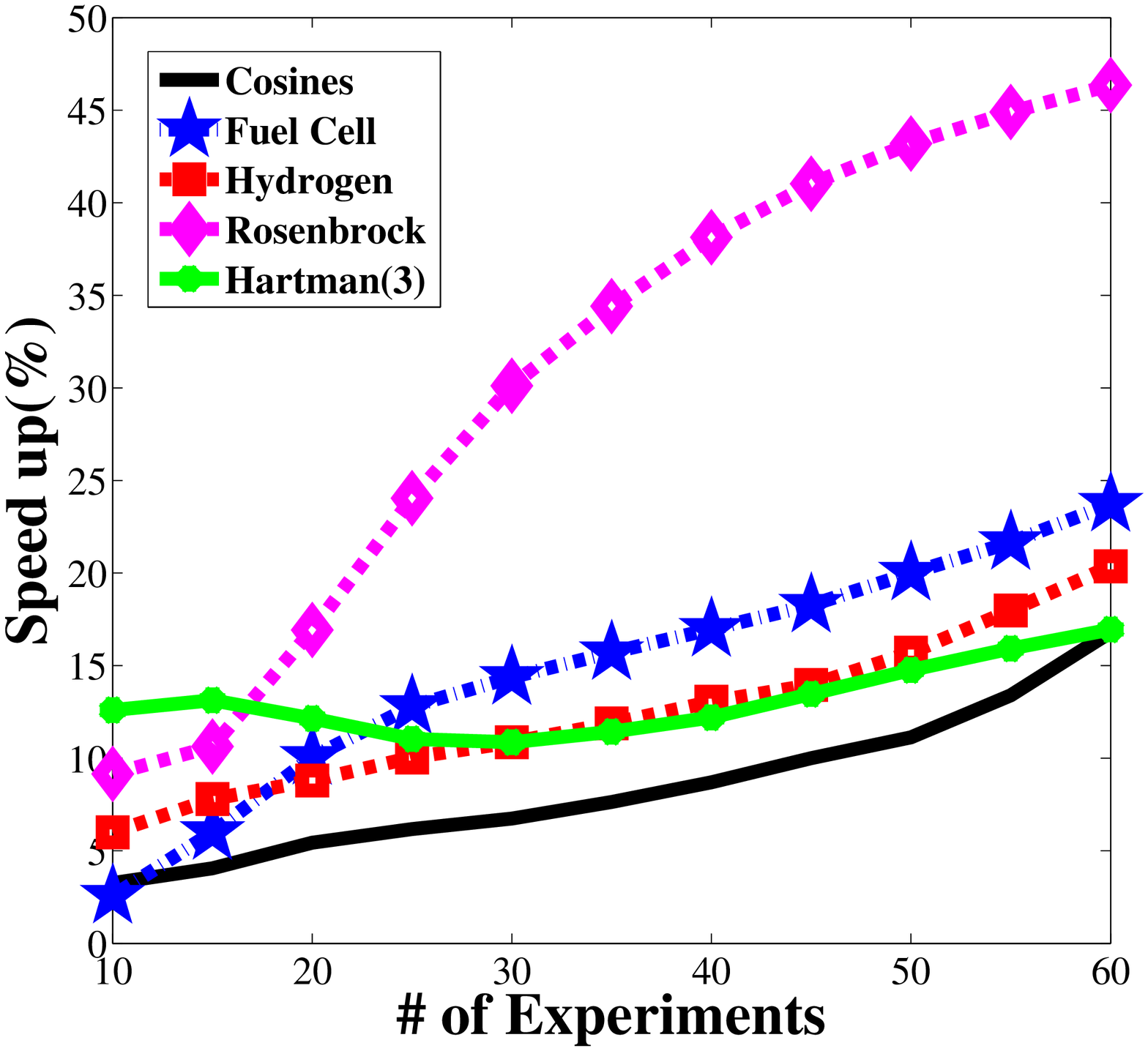}
\end{tabular}
\end{center} \vspace{-0.1in}
\label{fig:SvsB}
\end{figure}

Figure \ref{fig:SvsB} shows the speed up ratio versus the number of experiments in our $2,3$-dimensional benchmarks. It can be seen that as we increase the number of experiments the speedup ratio increases. Experimental results show that, when $n_l\geq 20$, we usually ask for $p_t>1$ points at each iteration and as we get closer to $60$, we usually reach the maximum possible batch size of $n_b=5$. The speedup ratio for Rosenbrock is significantly different from the other benchmarks since it has only one local/global maximum.

\bibliography{NIPS2011}
\bibliographystyle{abbrv}

\newpage

\appendix

\section{Proof of Theorem 1}
Recalling the notation introduced in the Theorem statement, we have
\begin{equation}
\begin{aligned}
\sigma_z^2 -\sigma^{*2}_z &= PA^{-1}P^T - [P\quad k^*_z]\left[
\begin{array}{cc}
A &  B^T \\B& D \end{array}\right]^{-1}\left[
\begin{array}{c}
P^T \\k^*_z \end{array}\right]\\
&= PA^{-1}P^T - [P\quad k^*_z]\left[\begin{array}{cc} A^{-1}+A^{-1}B^TmBA^{-1}& -A^{-1}B^Tm\\ -mBA^{-1}& m \end{array}\right]\left[
\begin{array}{c}
P^T \\k^*_z \end{array}\right]\\
&=\left(PA^{-1}B^T-k^*_z\right)m\left(BA^{-1}P^T-k^*_z\right).\\ 
\end{aligned}
\end{equation}
This concludes the proof of the theorem.\\

\section{Proof of Theorem 2}
Employing the notation of Theorem 1, we have
\begin{equation}
\begin{aligned}
\left|\mu_z-\mu^*_z\right| &= \left|PA^{-1}\boldsymbol{y}_{\mathcal{O}} - [P\quad k^*_z]\left[
\begin{array}{cc}
A &  B^T \\B& D \end{array}\right]^{-1}\left[
\begin{array}{c}
\boldsymbol{y}_{\mathcal{O}} \\\boldsymbol{y}^{*} \end{array}\right]\right|\\
&= \left|PA^{-1}\boldsymbol{y}_{\mathcal{O}} - [P\quad k^*_z]\left[
\begin{array}{cc}
A^{-1}+A^{-1}B^TmBA^{-1}& -A^{-1}B^Tm\\ -mBA^{-1}& m \end{array}\right]\left[
\begin{array}{c}
\boldsymbol{y}_{\mathcal{O}} \\\boldsymbol{y}^{*} \end{array}\right]\right|\\
&=\left|\left(PA^{-1}B^T-k^*_z\right)m\left(BA^{-1}\boldsymbol{y}_{\mathcal{O}}- \boldsymbol{y}^{*}\right)\right|.\\ 
\end{aligned}
\end{equation}

Notice that by Cauchy Shwarz inequality, for a fixed vector $v^T = \left(PA^{-1}B^T-k^*_z\right)m$ and random vector $u=BA^{-1}\boldsymbol{y}_{\mathcal{O}}-\boldsymbol{y}^{*}$, we have 
\begin{equation}
\begin{aligned}
E[|v^T\,u|] &\leq \left(\max_i |v_i|\right)E\left[\sum_i |u_i|\right]\\ &= \left(\max_i |v_i|\right)\sum_iE\left[|u_i|\right]\\ &=\left\|\left(PA^{-1}B^T-k^*_z\right)m\right\|_\infty  \sqrt{\frac{2}{\pi}}\|\boldsymbol{\sigma^{*}}\|_1.
\end{aligned}
\end{equation}

This concludes the proof of the Theorem.\\

\section{Proof of Theorem 3}

Taking the deriavtive with respect to $\mu$ and using the chain rule, we get
\begin{equation}
\begin{aligned}
\frac{d}{d\mu}\mbox{EI}(x)&=\frac{du}{d\mu}\frac{d}{du}\sigma_x\left(-u\Phi(-u)+ \phi(u)\right)\\
&=-\frac{1}{\sigma_x}\sigma_x\left(-\Phi(-u)+u\phi(-u)-u\phi(u)\right)\\
&=\Phi(-u)\geq 0.
\end{aligned}
\end{equation}

This concludes the proof of the Theorem.\\

\end{document}